\documentclass{article}

\usepackage[utf8]{inputenc}
\usepackage[T1]{fontenc}
\usepackage{amsmath,amssymb}
\usepackage{booktabs}
\usepackage{multirow}
\usepackage{graphicx}
\usepackage{hyperref}
\usepackage{natbib}
\usepackage{xcolor}
\usepackage{orcidlink}

\usepackage[margin=1in]{geometry}
\usepackage{setspace}
\newcommand{\eg}{\textit{e.g.}}

\title{Listen and Chant Before You Read:\\The Ladder of Beauty in LM Pre-Training}

\author{
  Yoshinori Nomura\,\orcidlink{0009-0004-4279-626X} \\
  Mirage Mountain Technologies Inc. \\
  \texttt{nomura@miragemt.com}
}

\date{}

\begin{document}

\maketitle

\begin{abstract}
We show that pre-training a Transformer on music before language
significantly accelerates language acquisition.
Using piano performances (MAESTRO dataset), a developmental
pipeline---music $\to$ poetry $\to$ prose---yields a 17.5\%
perplexity improvement over random initialization
($p < 0.001$, 5 seeds), with music and poetry improving
orthogonal model components (internal computation and embeddings,
respectively).
Convergence tests confirm that this is not a transient head start:
at $d\!=\!64$, multi-seed validation (5 seeds) shows a persistent
5.5\% gap at plateau ($p = 0.017$), with the pipeline converging
faster and to a lower loss in every run.
Real music matches the transfer ceiling of synthetic patterns
with one-third the data, and scaling experiments reveal that
optimal pre-training data volume shifts with model capacity
($-3\% \to +3\% \to +6\%$ advantage of larger datasets
from $d\!=\!16$ to $d\!=\!64$).
Across the scales we study ($d\!\in\!\{16,32,64\}$,
up to ${\sim}400$K parameters), these results suggest a
capacity-dependent data curation principle and indicate that
structured human creative outputs can provide an efficient
pre-training substrate for small language models; stronger
conclusions at modern pre-training scale will require
substantially larger experiments.
\end{abstract}

\section{Introduction}
\label{sec:intro}

Standard language model pre-training learns directly from text corpora.
Recent work has explored an additional stage \emph{before} this:
training first on non-linguistic data to establish general pattern
recognition capabilities, then proceeding to language.
We call this preliminary stage \emph{pre-pre-training}
(or \emph{foundation warming}), following \citet{lee2026nca},
who demonstrated that training on synthetic patterns generated
by Neural Cellular Automata (NCA)---two-dimensional discrete
dynamical systems that produce spatially structured patterns---improved
language model perplexity by up to 6\% and accelerated convergence
by 1.6$\times$.
This suggests that Transformers benefit from acquiring general
sequence modeling capabilities before encountering language.

However, NCA patterns are two-dimensional cellular automata
with limited structural similarity to language, which is inherently
a one-dimensional sequential structure.
We hypothesize that \emph{music}---a sequential, hierarchically
structured human creative output---provides a more natural
pre-pre-training substrate for language models.
The closest prior evidence for this comes from
\citet{papadimitriou2020learning}, who showed that an LSTM
pre-trained on MIDI music transfers positively to language
modeling on typologically diverse natural languages;
we revisit this hypothesis in a Transformer with causal
attention, compare real vs.\ synthetic music under a matched
tokenization, and embed the music stage in a staged developmental
pipeline (see \S\ref{sec:related}).

This hypothesis is grounded in cognitive science.
Human infants acquire sensitivity to pre-linguistic
regularities---rhythm, prosody, temporal patterns---before
lexical content \citep{kuhl2004early},
and extensive evidence suggests that music and language share
common cognitive substrates including hierarchical processing,
long-range dependency tracking, and expectation-based computation
\citep{patel2003language, koelsch2011toward}.
We argue that acquiring a primitive sense of regularity prior to
linguistic content is a fundamentally efficient strategy, and that
both biological and artificial learners converge on this solution:
music, as a rich source of pre-linguistic order, should
therefore be a superior pre-pre-training substrate.

We make six contributions:

\begin{enumerate}
    \item We demonstrate that music pre-pre-training accelerates
    language learning by 26\% in perplexity at the first epoch,
    persisting to 11.8\% after convergence
    ($p < 0.001$, 5 seeds).

    \item We show that data quality enables efficiency:
    real music by master composers reaches the same transfer
    ceiling as synthetic patterns with one-third the data volume.

    \item We discover that adding poetry as an intermediate phase
    yields additive improvements (17.5\% at epoch 2, $p < 0.001$),
    suggesting a developmental pipeline---music $\to$ poetry $\to$ prose---that
    parallels human language acquisition.

    \item We identify the mechanism:
    music improves internal computation (attention + FFN),
    while poetry calibrates token embeddings toward language.
    These orthogonal contributions explain the additive effect.

    \item We show that optimal pre-training data volume is a function
    of model capacity: the advantage of larger music datasets grows
    monotonically across three scales ($d\!=\!16$, $32$, $64$),
    revealing a capacity-dependent data curation principle.

    \item We verify via convergence tests that the pipeline's
    advantage is not a transient head start: multi-seed validation
    at $d\!=\!64$ shows a persistent 5.5\% gap at plateau
    ($p = 0.017$), with the pipeline converging faster and to a
    lower loss in every run.
\end{enumerate}

\section{Related Work}
\label{sec:related}

\paragraph{Pre-pre-training and foundation warming.}
The idea of training on auxiliary data before the main pre-training
phase has emerged recently.
\citet{lee2026nca} introduced Neural Cellular Automata (NCA) as
a source of synthetic pre-pre-training data,
showing that non-linguistic pattern recognition transfers to
language modeling.
Their ablation study revealed that attention weights are
the most transferable component,
while MLP layers encode domain-specific statistics that can
interfere with downstream learning.
Our work extends this line of research by replacing synthetic 2D
patterns with real 1D musical sequences, achieving substantially
stronger transfer effects.

\paragraph{Music and Transformers.}
Music Transformer \citep{huang2018music} demonstrated that
self-attention can model long-range musical structure,
and subsequent work has developed sophisticated MIDI tokenization
schemes \citep{zeng2021musicbert}.
These works focus on music generation rather than
cross-domain transfer.

\paragraph{Music-to-language transfer.}
The most directly related prior work is
\citet{papadimitriou2020learning}, who showed that pre-training an
LSTM on MIDI music improves subsequent language modeling on
typologically diverse natural languages, and used this
transfer effect as a probe to argue that hierarchical structure
is what is being acquired and reused across modalities.
Our work builds on this line in several ways:
(i) we instantiate the idea in a Transformer with causal
attention, rather than an LSTM;
(ii) we compare real (MAESTRO) against rule-based synthetic
music under a matched tokenization, isolating the contribution
of ``musical quality'' beyond mere structured non-linguistic
sequences;
(iii) we introduce an intermediate poetry stage, yielding a
three-phase music~$\to$~poetry~$\to$~prose pipeline whose
components we show to act on orthogonal model parts
(internal computation vs.\ embeddings); and
(iv) we characterize how optimal pre-training data volume
scales with model capacity.
We therefore do not claim that music-to-language transfer is
novel; our contributions concern the mechanism, the staged
developmental pipeline, and the capacity--data interaction.

\paragraph{Cross-modal transfer learning (other modalities).}
Transfer between modalities has been studied primarily in the
vision-language direction \citep{lu2019vilbert}.
Independently of the music-to-language line above, there is a
long tradition in cognitive science of treating music and language
as sharing processing resources
\citep{patel2003language, koelsch2011toward},
which we translate into a concrete training methodology.

\paragraph{Music and language in cognitive science.}
\citet{patel2003language} proposed the Shared Syntactic Integration
Resource Hypothesis (SSIRH), arguing that music and language
share neural resources for processing hierarchical structure.
\citet{koelsch2011toward} documented overlapping neural substrates
for processing expectations and violations in both domains.
\citet{kuhl2004early} showed that infants acquire sensitivity to
pre-linguistic regularities---rhythm, prosody, temporal patterns---before
lexical content, consistent with the view that acquiring a primitive
sense of regularity prior to linguistic content is a fundamentally
efficient strategy that both biological and artificial learners
converge on.

\section{Method}
\label{sec:method}

\subsection{Model}
\label{sec:model}

We use a standard autoregressive Transformer decoder
\citep[GPT-2 architecture;][]{radford2019language} with causal (left-to-right)
self-attention and learned absolute position embeddings.
We deliberately choose small dimensions to study learning
efficiency under capacity constraints:
$d_\text{model} = 16$ (the internal representation width),
$n_\text{heads} = 1$ (single attention head),
$d_\text{head} = 16$ (per-head dimension, equal to $d_\text{model}$),
$n_\text{layers} = 8$ (Transformer blocks),
$d_\text{ff} = 64$ (feedforward hidden dimension, $4 \times d_\text{model}$),
yielding approximately 33K trainable parameters.
We verify our findings at larger scales
($d = 32$, $d = 64$) in Section~\ref{sec:scale}.

\subsection{Music Tokenization}
\label{sec:tokenization}

MIDI (Musical Instrument Digital Interface) files encode musical
performances as sequences of discrete note events, each specified
by pitch, onset time, duration, and velocity (loudness).
We convert these events into a flat token sequence using a simplified
version of the REMI (REvamped MIDI-derived) tokenization
\citep{huang2020pop}, which represents music as a linear sequence
of tokens organized bar by bar.

Our vocabulary consists of 160 tokens in five categories:

\begin{itemize}
    \item \textbf{Special tokens} (4): PAD (padding), BOS (beginning of
    sequence), EOS (end of sequence), BAR (bar boundary marker)
    \item \textbf{Position} (16): position within a bar on a 16th-note
    grid (0--15), indicating \emph{when} a note occurs within the bar
    \item \textbf{Pitch} (128): MIDI pitch values (0--127, where 60 =
    middle C), indicating \emph{which} note is played
    \item \textbf{Duration} (8): note length in 16th-note units (1--8),
    indicating \emph{how long} the note sustains
    \item \textbf{Velocity} (4): dynamics bins (pp, p, f, ff),
    indicating \emph{how loudly} the note is played
\end{itemize}

Each note event is represented by exactly four tokens
(Position, Pitch, Duration, Velocity) in a fixed order,
producing a deterministic \emph{token grammar}---a set of strict
ordering constraints on which token types can follow which:
$\text{BAR} \to \text{POS} \to \text{PITCH} \to \text{DUR} \to \text{VEL}
\to (\text{POS} \mid \text{BAR} \mid \text{EOS})$.
This grammar means the model must learn both the local syntax
(the fixed ordering of token types within a note) and the musical
content (which pitches, durations, and dynamics to predict).

\subsection{Datasets}
\label{sec:data}

All sequential data is divided into \emph{chunks}: fixed-length
subsequences of $\text{seq\_len} + 1 = 257$ tokens, where the first
256 tokens serve as input and the last 256 as the prediction target
(shifted by one position).
Chunks are created by concatenating all source material and splitting
into non-overlapping windows.
The number of chunks thus determines the effective dataset size.

\begin{table}[h]
\centering
\caption{Datasets used in our experiments.}
\label{tab:datasets}
\begin{tabular}{@{}llrl@{}}
\toprule
Dataset & Source & Chunks & Nature \\
\midrule
Synthetic music & Algorithmic generation & varied & Rule-based patterns \\
MAESTRO v2 & Piano performances & 36,061 & 58 composers, master pianists \\
Gutenberg Poetry & Project Gutenberg & 36,000 & Classical English poetry \\
WikiText-103 & Wikipedia & --- & General English prose \\
\bottomrule
\end{tabular}
\end{table}

\paragraph{MAESTRO.}
The MAESTRO dataset \citep{hawthorne2019enabling} contains
aligned MIDI and audio from piano performances by
professional pianists, covering composers from Bach to
Rachmaninoff (58 composers, 1,276 pieces).
We use the note-level annotations
(pitch, onset time, offset time, velocity)
and tokenize directly without relying on MIDI files.

\paragraph{Synthetic music.}
As a baseline data source, we generate synthetic music via a
rule-based algorithm using the same MIDI token vocabulary as MAESTRO.
For each piece, the generator (i) selects a random root note
(C3--C5) and scale (major, minor, or pentatonic),
(ii) creates a short motif of 1--2 bars by placing 2--6 notes
at random positions within a 16th-note grid, and
(iii) extends the piece to 4--16 bars by sampling from four
operations: exact repetition (40\%), transposition by a
diatonic interval (20\%), pitch variation where 30\% of notes
are replaced with scale-compatible alternatives (25\%),
or generation of a new contrasting phrase (15\%).
The resulting sequences exhibit surface-level musical structure---repetition,
scale-constrained pitch, and simple variation---but lack the
harmonic progressions, voice leading, phrase-level tension--resolution
arcs, and long-range structural planning found in composed music.
This design is intentionally minimal: it serves as a controlled
baseline to isolate the contribution of \emph{musical quality}
(i.e., the structural richness present in real performances)
from mere exposure to structured non-linguistic sequences.

\paragraph{Gutenberg Poetry Corpus.}
The corpus \citep{biglam2022gutenberg} contains 3 million
lines of English poetry from Project Gutenberg, tokenized
with the GPT-2 tokenizer (vocabulary size 50,257).
We subsample to 36,000 chunks to match the MAESTRO data size.

\paragraph{WikiText-103.}
We use a 10\% subsample of WikiText-103 \citep{merity2017pointer}
for language evaluation, tokenized with the GPT-2 tokenizer.

\subsection{Training Pipeline}
\label{sec:pipeline}

\begin{figure*}[t]
\centering
\includegraphics[width=\textwidth]{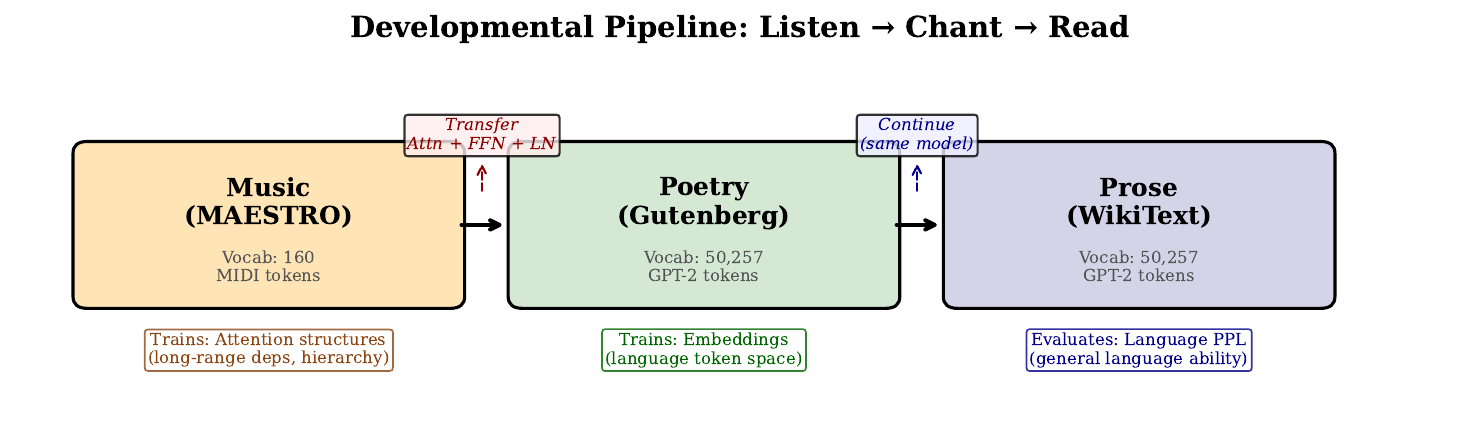}
\caption{The developmental pipeline.
Music pre-training establishes attention structures
(long-range dependency tracking, hierarchical pattern recognition);
poetry calibrates token embeddings toward the language space;
prose training evaluates general language modeling ability.
The vocabulary change between music (160 tokens) and poetry/prose
(50,257 tokens) requires selective weight transfer
(Section~\ref{sec:pipeline}).}
\label{fig:pipeline}
\end{figure*}

We define a \emph{developmental pipeline} (Figure~\ref{fig:pipeline}):
a sequence of training phases on progressively more language-like data,
where each phase builds on the representations learned in the previous one.
The term ``developmental'' is inspired by the trajectory of human
language acquisition, which progresses from sub-linguistic
pattern recognition (rhythm, prosody) through structured language
(nursery rhymes, songs) to general prose comprehension.

Our pipeline consists of three phases:

\begin{enumerate}
    \item \textbf{Music phase}: Train a Transformer on music tokens
    (vocabulary size 160) until convergence.
    Training runs for up to 200 epochs with \emph{early stopping}:
    if validation loss does not improve for 20 consecutive epochs
    (the \emph{patience} parameter), training halts and the
    best checkpoint is retained.

    \item \textbf{Poetry phase}: Construct a new model with
    a language vocabulary (GPT-2 tokenizer, 50,257 tokens) and
    transfer the learned weights via \emph{selective weight transfer}
    (described below).
    Train on poetry for 3 epochs.

    \item \textbf{Prose phase}: Continue training the same model
    on WikiText-103 for 3 epochs.
    Evaluate using \emph{perplexity} (PPL), defined as
    $\text{PPL} = \exp(\mathcal{L})$ where $\mathcal{L}$ is the
    mean cross-entropy loss per token on the validation set.
    Lower perplexity indicates better language modeling:
    intuitively, it measures how many tokens the model is
    ``choosing among'' at each prediction step.
\end{enumerate}

\paragraph{Selective weight transfer.}
When transitioning between phases with different vocabularies
(music $\to$ poetry, or music $\to$ prose), we face a vocabulary
mismatch: the music model uses 160 tokens while the language model
uses 50,257.
Our solution is to partition the model's parameters into two groups:

\begin{itemize}
    \item \textbf{Internal computation layers}---attention weights
    ($W_Q$, $W_K$, $W_V$, $W_O$), feedforward weights ($W_1$, $W_2$),
    and layer normalization parameters ($\gamma$, $\beta$)---are
    \emph{transferred}.
    These layers implement domain-general computation:
    dependency tracking, pattern recognition, and representation
    transformation.
    Their dimensions depend only on $d_\text{model}$ and are
    identical across vocabularies.

    \item \textbf{Vocabulary-dependent layers}---the token embedding
    matrix and the output projection (language model head)---are
    \emph{reinitialized} randomly.
    These layers map between the token space and the internal
    representation space, and their dimensions depend on vocabulary
    size.
    Since music tokens (note events) and language tokens (subwords)
    have no correspondence, these weights cannot be meaningfully
    transferred.
\end{itemize}

This design is motivated by \citet{lee2026nca}'s finding that
attention weights are the primary carriers of transferable
computation, while MLP weights encode domain-specific statistics.
A practical consequence is that immediately after transfer,
the model's perplexity is near-random (${\sim}50{,}000$) because
the embeddings carry no information;
the transferred internal weights accelerate \emph{how quickly}
the model learns, not where it starts.

\subsection{Experimental Design}
\label{sec:conditions}

Our experiments are organized in three phases, each addressing
a distinct question.

\subsubsection{Phase 1: Data Volume Control (Apples-to-Apples Comparison)}
\label{sec:phase1}

A na\"ive comparison of synthetic music (2,881 chunks) against
MAESTRO (36,061 chunks) confounds data \emph{quality} with data
\emph{quantity}.
To disentangle the two, we control data volume at three levels
(3k, 12k, and 36k chunks) and compare synthetic vs.\ MAESTRO
at each level.
This yields six pre-training conditions plus a random
initialization baseline:

\begin{table}[h]
\centering
\caption{Phase 1: Data volume-controlled conditions. Each condition
trains a $d\!=\!16$ music model until convergence (patience 20),
then transfers internal weights to a language model and evaluates on
WikiText-103 (10\% subsample) for 3 epochs.}
\label{tab:phase1-conditions}
\begin{tabular}{@{}llrl@{}}
\toprule
Condition & Source & Chunks & Purpose \\
\midrule
Random & --- & --- & Baseline (no music pre-training) \\
Synth-3k & Synthetic & 3,000 & \multirow{3}{*}{Quality effect at matched volume} \\
Synth-12k & Synthetic & 12,000 & \\
Synth-36k & Synthetic & 36,000 & \\
MAESTRO-3k & MAESTRO subsample & 3,000 & \multirow{3}{*}{Quality effect at matched volume} \\
MAESTRO-12k & MAESTRO subsample & 12,000 & \\
MAESTRO-36k & MAESTRO (full) & 36,000 & \\
\bottomrule
\end{tabular}
\end{table}

If MAESTRO outperforms synthetic data at matched volume,
we can attribute the difference to data quality---the structural
richness of composed music---rather than quantity.
A secondary question is how pre-training data volume interacts
with model capacity: does the optimal amount saturate
at a level determined by the model's ability to absorb structure?

\subsubsection{Phase 2: Statistical Validation (Multi-Seed)}
\label{sec:phase2}

Phase 1 uses a single seed ($\text{seed}=42$).
To establish statistical significance, we re-run four key conditions
across five seeds (42, 123, 456, 789, 1024):

\begin{table}[h]
\centering
\caption{Phase 2: Multi-seed conditions. Music checkpoints are
shared (seed 42); only the language learning phase varies by seed.}
\label{tab:phase2-conditions}
\begin{tabular}{@{}lll@{}}
\toprule
Condition & Pipeline & Purpose \\
\midrule
A & Random $\to$ Prose & Baseline \\
B & MAESTRO-12k $\to$ Prose & Best single-source from Phase 1 \\
C & MAESTRO-36k $\to$ Poetry $\to$ Prose & Full developmental pipeline \\
D & Synth-36k $\to$ Prose & Data quality control \\
\bottomrule
\end{tabular}
\end{table}

Music checkpoints are fixed across seeds: the stochasticity
that matters is in language learning (random initialization of
embeddings and LM head, data shuffling).
We report mean $\pm$ std and paired $t$-tests for each condition
vs.\ random baseline.

Note that condition B uses MAESTRO-12k rather than MAESTRO-36k.
This choice is informed by Phase 1's finding that MAESTRO-12k
achieves the best transfer at $d\!=\!16$ (Section~\ref{sec:quality}),
suggesting that the small model saturates before exhausting
the full 36k dataset.
Condition C retains MAESTRO-36k for the music phase because the
poetry phase provides an additional pathway for the model to
exploit the richer pre-training.

\paragraph{Compute-matched control.}
The developmental pipeline (condition C) involves more total
training steps than the baseline: 3 epochs of poetry
(${\sim}6{,}075$ batches) followed by 3 epochs of prose
(${\sim}8{,}526$ batches), totaling ${\sim}14{,}600$ batches,
versus ${\sim}8{,}500$ batches for the 3-epoch prose-only conditions.
To rule out the possibility that the pipeline's advantage
is simply due to additional compute, we include a
\emph{compute-matched} control: random initialization trained
on WikiText-103 for 5 epochs (${\sim}14{,}210$ batches),
matching the pipeline's total training budget to within 3\%.

\subsubsection{Phase 3: Scale $\times$ Data-Size Interaction}
\label{sec:phase3-design}

Phase 1 revealed a surprising result: at $d\!=\!16$, MAESTRO-12k
\emph{outperforms} MAESTRO-36k.
We hypothesize that this reflects model capacity saturation---the
33K-parameter model cannot absorb more structure than 12k chunks
provide.
If so, larger models should shift the optimal data size upward.

To test this, we run three conditions (random, MAESTRO-12k,
MAESTRO-36k) at three model scales:

\begin{table}[h]
\centering
\caption{Phase 3: Scale experiment design. All models use 8 layers;
$d\!=\!16$ reuses Phase 1 checkpoints.}
\label{tab:phase3-conditions}
\begin{tabular}{@{}rrrrl@{}}
\toprule
$d_\text{model}$ & Heads & $d_\text{ff}$ & Params & Key question \\
\midrule
16 & 1 & 64 & 33K & Phase 1 reuse (12k $>$ 36k) \\
32 & 2 & 128 & 130K & Does 36k begin to overtake 12k? \\
64 & 4 & 256 & 400K & Does the reversal strengthen? \\
\bottomrule
\end{tabular}
\end{table}

This design tests whether optimal pre-training data volume scales
with model capacity---a question with practical implications for
data curation at larger scales.
\citet{lee2026nca} showed that the optimal \emph{complexity} of
NCA patterns varies by target domain (low complexity for code,
high for math).
Our experiment extends this finding to a second axis: not just
\emph{what kind} of data to use, but \emph{how much},
as a function of the model's capacity to absorb it.

\subsubsection{Shared Hyperparameters}

All conditions across all phases use identical hyperparameters:
learning rate $10^{-3}$ with cosine decay to $10^{-4}$,
200-step linear warmup, AdamW optimizer with weight decay 0.1,
gradient clipping at 1.0, and gradient accumulation over 2 steps.
Music models train for up to 200 epochs with early stopping
(patience 20).
Language evaluation uses 3 epochs on a 10\% subsample of
WikiText-103.
The poetry phase, when present, consists of 3 epochs on
36,000 chunks of Gutenberg poetry.
Phases 1 and 3 use a fixed random seed ($\text{seed}=42$);
Phase 2 varies the seed across five values
(42, 123, 456, 789, 1024) for statistical validation.

\section{Results}
\label{sec:results}

\subsection{What the Music Model Learns}
\label{sec:music-learns}

Before examining transfer effects, we characterize what the
$d=16$ model learns from music data.

\paragraph{Token grammar.}
The model learns the deterministic token grammar nearly perfectly:
after BAR, it predicts POS with 97.8\% probability;
after POS, PITCH with 99.7\%; after PITCH, DUR with 99.9\%;
after DUR, VEL with 99.8\%.

\paragraph{Pattern completion.}
When presented with a motif (C--E--G) repeated three times,
the model predicts BAR (pattern continuation) with 70.2\%
probability, and after BAR, predicts the correct starting
position with 89.4\% probability.
This demonstrates genuine pattern recognition beyond
token-level statistics.

\paragraph{Attention specialization.}
The single attention head devotes 89.5\% of its attention mass
to positions more than 8 tokens away,
indicating specialization for long-range dependency tracking.
Local patterns (the token grammar) are handled by the
feedforward layers and embeddings.

\subsection{Phase 1: Data Quality at Controlled Volume}
\label{sec:quality}

Table~\ref{tab:phase1-results} shows the Phase 1 results:
WikiText-103 perplexity after 3 epochs of language learning,
for each pre-training condition at matched data volumes.

\begin{table}[h]
\centering
\caption{Phase 1 results: WikiText-103 validation perplexity
at epoch 2 (final). Percentage shows improvement vs.\ random baseline.}
\label{tab:phase1-results}
\begin{tabular}{@{}lrrr@{}}
\toprule
Condition & E0 & E1 & E2 \\
\midrule
Random (baseline) & 695.1 & 484.7 & 421.5 \\
\midrule
Synth-3k   & 611.6 & 456.4 & 401.5 ($-$4.7\%) \\
Synth-12k  & 525.1 & 418.8 & 383.9 ($-$8.9\%) \\
Synth-36k  & 494.3 & 398.3 & 370.7 ($-$12.1\%) \\
\midrule
MAESTRO-3k  & 603.5 & 452.4 & 403.9 ($-$4.2\%) \\
MAESTRO-12k & 514.7 & 404.7 & \textbf{366.7} ($-$\textbf{13.0\%}) \\
MAESTRO-36k & 503.8 & 404.7 & 375.4 ($-$10.9\%) \\
\bottomrule
\end{tabular}
\end{table}

Three findings emerge.
First, all pre-training conditions outperform the random baseline,
confirming that music pre-pre-training is beneficial regardless
of data source or volume.

Second, \textbf{MAESTRO-12k achieves the best overall transfer}
(PPL 366.7), surpassing even Synth-36k (PPL 370.7) with one-third
the data volume.
This provides direct evidence that data \emph{quality}---the
structural richness of real performances---can compensate for
data quantity.

Third, \textbf{MAESTRO shows non-monotonic scaling}: performance
improves from 3k to 12k but \emph{degrades} from 12k to 36k
(366.7 $\to$ 375.4).
Synthetic data, by contrast, improves monotonically
(401.5 $\to$ 383.9 $\to$ 370.7).
We interpret this as capacity saturation: the 33K-parameter
model can absorb the structural patterns in approximately
12k chunks of real music; additional data introduces redundancy
or noise that slightly degrades transfer.
Synthetic data, being structurally simpler, continues to
benefit from additional examples because the model has not yet
extracted all learnable patterns.
Phase 3 (Section~\ref{sec:scale}) tests this hypothesis by
examining whether larger models shift the optimal data volume
upward.

\subsection{Phase 2: Statistical Validation}
\label{sec:transfer}

Table~\ref{tab:phase2-results} reports the multi-seed results.
All four pre-training conditions are evaluated across five random
seeds; the music checkpoints are shared (seed 42) and only the
language learning phase varies.

\begin{table}[h]
\centering
\caption{Phase 2 results: WikiText-103 validation perplexity
(mean $\pm$ std over 5 seeds).
$\Delta$ is improvement vs.\ random baseline at epoch 2.
All conditions vs.\ random are significant at $p < 0.001$
(paired $t$-test, $n=5$).}
\label{tab:phase2-results}
\begin{tabular}{@{}lrrrrl@{}}
\toprule
Condition & E0 & E1 & E2 & $\Delta$ & $p$ \\
\midrule
Random (baseline)
  & $694.1 \pm 17.6$
  & $483.4 \pm 7.7$
  & $423.0 \pm 5.3$
  & ---
  & --- \\
MAESTRO-12k $\to$ Prose
  & $512.5 \pm 2.5$
  & $407.3 \pm 2.4$
  & $373.2 \pm 3.8$
  & $-11.8\%$
  & $< 0.001$ \\
Synth-36k $\to$ Prose
  & $499.4 \pm 6.2$
  & $402.5 \pm 4.7$
  & $371.5 \pm 3.5$
  & $-12.2\%$
  & $< 0.001$ \\
MAESTRO $\to$ Poetry $\to$ Prose
  & $415.7 \pm 4.4$
  & $369.5 \pm 4.2$
  & $\mathbf{349.0 \pm 5.8}$
  & $\mathbf{-17.5\%}$
  & $< 0.001$ \\
\bottomrule
\end{tabular}
\end{table}

Three results stand out.

First, \textbf{the findings from Phase 1 replicate with high
consistency}.
Standard deviations are small (3--6 PPL points), and all three
pre-training conditions significantly outperform the random baseline
($p < 0.001$ for all comparisons, paired $t$-test, $n=5$).

Second, \textbf{MAESTRO-12k and Synth-36k are statistically
indistinguishable} at epoch 2 (373.2 vs.\ 371.5; $t = 0.97$,
$p = 0.39$).
This is notable: MAESTRO achieves comparable performance with
one-third the data volume, but the advantage is not statistically
significant at this sample size.
The qualitative advantage of real music identified in Phase 1
(Section~\ref{sec:quality}) is thus better characterized as
an \emph{efficiency} advantage---fewer data points needed to
reach the same transfer effect---rather than a ceiling advantage.

Third, \textbf{the developmental pipeline achieves the strongest
transfer by a wide margin}.
The MAESTRO $\to$ Poetry $\to$ Prose condition (PPL 349.0)
outperforms the next-best condition (Synth-36k, PPL 371.5) by
6.1\%, and this difference is highly significant
(vs.\ MAESTRO-12k: $t = 10.86$, $p < 0.001$).
The poetry phase provides an additional 5.7 percentage points
of improvement beyond music alone (17.5\% vs.\ 11.8\% relative
to baseline), and this increment is consistent across all five
seeds.

The epoch-0 perplexity reveals the mechanism behind the poetry
phase's advantage.
The random baseline starts at PPL $694 \pm 18$;
direct-transfer conditions (MAESTRO-12k, Synth-36k) start at
${\sim}500$---lower but still high, because the reinitialized
embeddings carry no language-specific information.
The poetry pipeline starts at $416 \pm 4$, already below the
random baseline's \emph{final} performance after three full
epochs of language learning.
This dramatic head start reflects the fact that the poetry phase
has already adjusted the embeddings toward the language token space,
so the model enters the prose phase with both trained internal
weights \emph{and} partially calibrated embeddings.

\paragraph{Compute-matched control.}
The developmental pipeline uses more total training steps than
the 3-epoch prose-only conditions (${\sim}14{,}600$ vs.\
${\sim}8{,}500$ batches).
To rule out that the pipeline's advantage is simply due to
additional compute, we train a randomly initialized model on
WikiText-103 for 5 epochs (${\sim}14{,}210$ batches), matching
the pipeline's total budget to within 3\%.
Table~\ref{tab:compute-matched} shows the results.

\begin{table}[h]
\centering
\caption{Compute-matched control: WikiText-103 validation perplexity
(mean $\pm$ std over 5 seeds).
The compute-matched baseline trains for 5 prose epochs
(${\sim}14{,}210$ batches) vs.\ the pipeline's ${\sim}14{,}600$
batches (music + poetry + prose).}
\label{tab:compute-matched}
\begin{tabular}{@{}lrrl@{}}
\toprule
Condition & Batches & Final PPL & $p$ (vs.\ pipeline) \\
\midrule
Random (3 ep prose)
  & ${\sim}8{,}500$
  & $423.0 \pm 5.3$
  & --- \\
Compute-matched (5 ep prose)
  & ${\sim}14{,}210$
  & $367.3 \pm 2.3$
  & --- \\
MAESTRO $\to$ Poetry $\to$ Prose
  & ${\sim}14{,}600$
  & $\mathbf{349.0 \pm 5.8}$
  & $0.005$ \\
\bottomrule
\end{tabular}
\end{table}

The compute-matched baseline (PPL $367.3 \pm 2.3$) substantially
outperforms the 3-epoch random baseline ($423.0$), confirming that
additional training helps.
However, the developmental pipeline (PPL $349.0 \pm 5.8$) still
significantly outperforms the compute-matched control
($t = 5.47$, $p = 0.005$, paired $t$-test, $n=5$).
The pipeline's 17.5\% improvement over the 3-epoch baseline cannot
be attributed to additional compute alone:
at matched compute budget, the pipeline retains a 5.0\% advantage
over pure prose training, demonstrating that the \emph{content}
of the developmental stages---not merely their duration---drives
the transfer effect.

\subsection{Phase 3: Scale $\times$ Data-Size Interaction}
\label{sec:scale}

Table~\ref{tab:phase3-results} shows the Phase 3 results:
WikiText-103 perplexity at epoch 2 for each condition at three
model scales.

\begin{table}[h]
\centering
\caption{Phase 3 results: WikiText-103 validation perplexity at epoch 2.
$\Delta_R$ is improvement vs.\ random baseline at the same scale.
$\Delta_{12/36}$ compares 36k against 12k (positive = 36k is better).}
\label{tab:phase3-results}
\begin{tabular}{@{}lrrrrr@{}}
\toprule
Scale & Random & MAESTRO-12k ($\Delta_R$) & MAESTRO-36k ($\Delta_R$) & $\Delta_{12/36}$ \\
\midrule
$d\!=\!16$ (33K)  & 418.9 & 364.3 ($-$13.0\%) & 375.4 ($-$10.4\%) & $-$3.1\% \\
$d\!=\!32$ (130K) & 263.3 & 222.2 ($-$15.6\%) & 215.0 ($-$18.4\%) & $+$3.3\% \\
$d\!=\!64$ (400K) & 167.2 & 149.1 ($-$10.8\%) & \textbf{140.0} ($-$\textbf{16.3\%}) & $+$\textbf{6.1\%} \\
\bottomrule
\end{tabular}
\end{table}

The $d\!=\!16$ row differs slightly from Table~\ref{tab:phase1-results}
because Phase 3 re-runs the language transfer on a different machine;
the qualitative pattern (12k $>$ 36k) is identical.\footnote{%
Phase 1 random baseline: 421.5; Phase 3: 418.9.
The difference ($<1\%$) reflects numerical non-determinism
across hardware and CUDA versions.}

The results confirm the capacity saturation hypothesis and reveal
a striking monotonic trend.
At $d\!=\!16$, MAESTRO-12k outperforms 36k by 3.1\%---the model
is too small to absorb the additional structure in the larger dataset.
At $d\!=\!32$, the relationship reverses: 36k overtakes 12k by 3.3\%.
At $d\!=\!64$, the advantage of 36k widens further to 6.1\%.

\begin{figure*}[t]
\centering
\includegraphics[width=\textwidth]{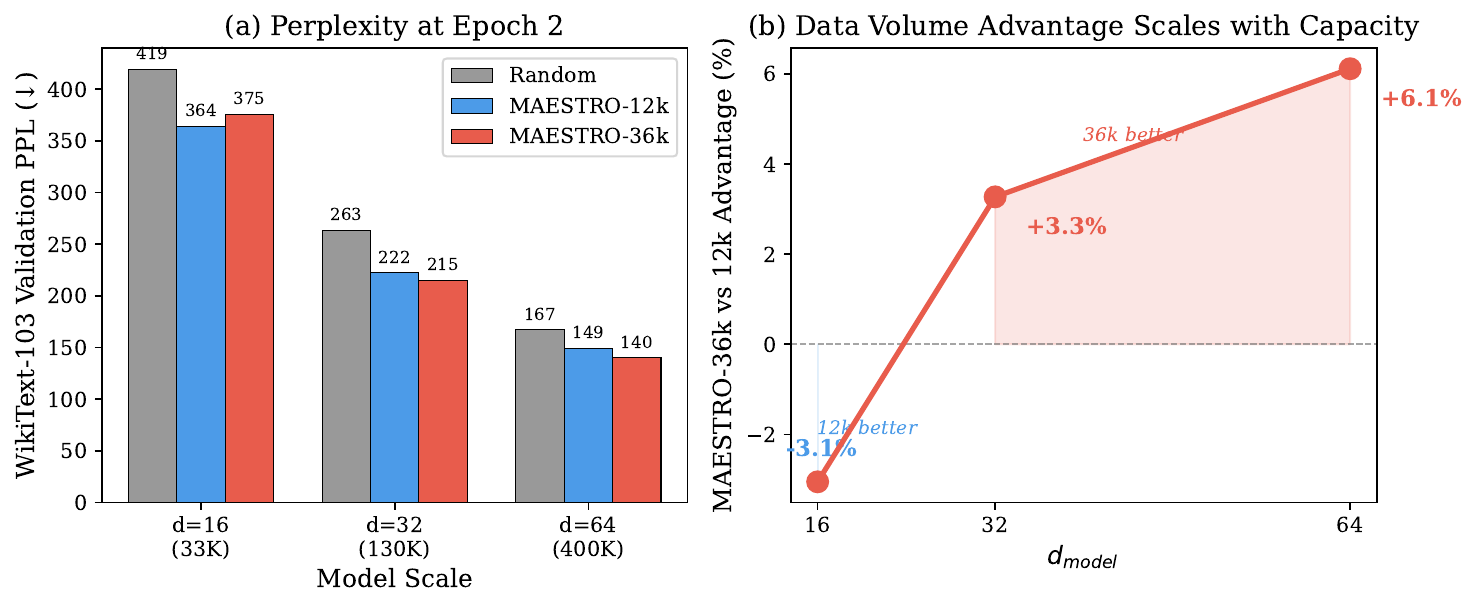}
\caption{Phase 3: Scale $\times$ data-size interaction.
\textbf{(a)} WikiText-103 validation perplexity at epoch 2 across
three model scales.
Music pre-training consistently improves over the random baseline
at all scales, and the improvement grows with model size.
\textbf{(b)} The advantage of MAESTRO-36k over MAESTRO-12k
increases monotonically with model capacity
($-3.1\% \to +3.3\% \to +6.1\%$),
confirming the capacity saturation hypothesis:
small models cannot absorb large datasets,
but larger models increasingly benefit from more data.}
\label{fig:scale}
\end{figure*}

\begin{figure*}[t]
\centering
\includegraphics[width=\textwidth]{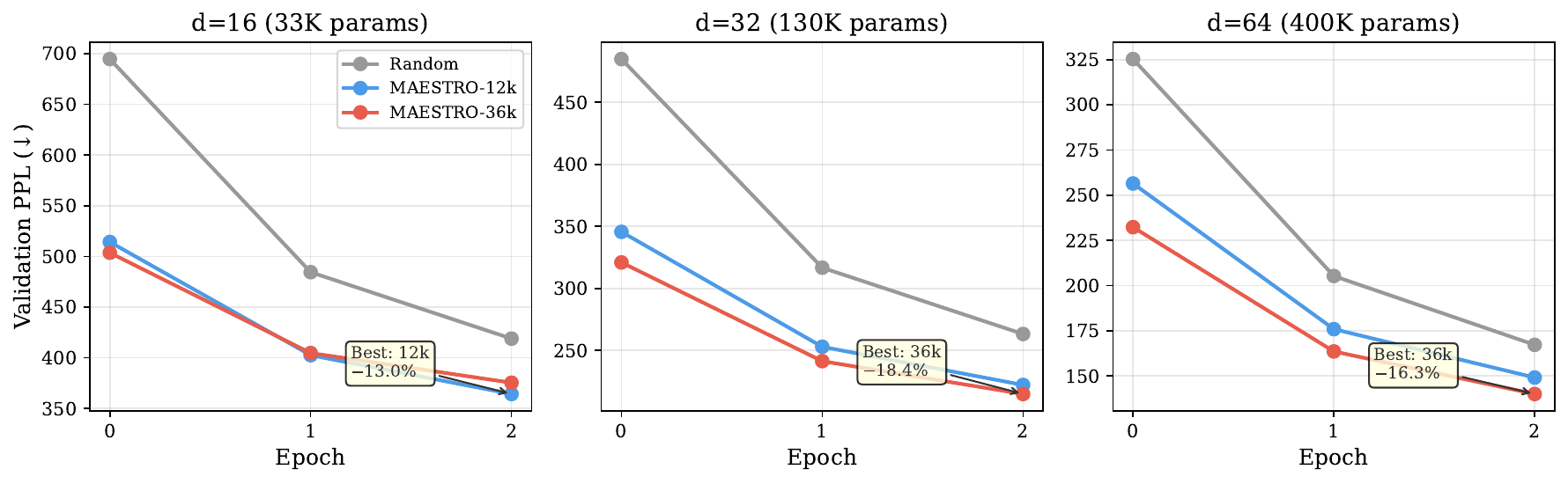}
\caption{Learning curves across scales.
At $d\!=\!16$, MAESTRO-12k (blue) is the best condition;
at $d\!=\!32$ and $d\!=\!64$, MAESTRO-36k (red) overtakes 12k,
with the gap widening at larger scales.
The annotation shows the best music condition and its improvement
over the random baseline at epoch 2.}
\label{fig:curves}
\end{figure*}

Figure~\ref{fig:scale} visualizes these results, and
Figure~\ref{fig:curves} shows the corresponding learning curves.
The trajectory of the $\Delta_{12/36}$ column ($-3.1\% \to +3.3\%
\to +6.1\%$) is monotonically increasing, indicating that the
advantage of larger pre-training datasets grows systematically
with model capacity.
Meanwhile, the 12k condition shows diminishing returns at larger
scales ($-13.0\% \to -15.6\% \to -10.8\%$), consistent with
insufficient data for the model's capacity.
The 36k condition, by contrast, maintains strong and stable
transfer ($-10.4\% \to -18.4\% \to -16.3\%$), and the
continued expansion of the 36k--12k gap at $d\!=\!64$ suggests
that even 36k chunks may be insufficient at this scale---the
optimal data volume likely exceeds 36k for 400K-parameter models.

These results extend \citeauthor{lee2026nca}'s finding that
the optimal \emph{complexity} of pre-training data varies by
target domain to a second axis: the optimal \emph{volume} of
pre-training data varies by model capacity.

\subsection{Adding the Poetry Phase at $d\!=\!32$}
\label{sec:phase3-full}

Phase 3 (Table~\ref{tab:phase3-results}) evaluates the music-only
transfer (MAESTRO $\to$ prose) across scales.
Phase 2 establishes the full developmental pipeline
(music $\to$ poetry $\to$ prose) at $d\!=\!16$, but the full pipeline
had not been evaluated at $d\!=\!32$.
To fill this gap, we run the full pipeline at $d\!=\!32$ using the
Phase 3 protocol (seed 42, WikiText-103 epoch 2 evaluation):

\begin{table}[h]
\centering
\caption{Full developmental pipeline at $d\!=\!32$
(seed 42, WikiText-103 validation PPL at epoch 2). The
Poetry phase contributes an additional $-8.7\%$ over the
music-only condition.}
\label{tab:phase3-full-d32}
\begin{tabular}{@{}lcc@{}}
\toprule
Condition ($d\!=\!32$) & E2 PPL & $\Delta$ vs random \\
\midrule
Random                                    & 263.3 & --- \\
MAESTRO-36k $\to$ prose                   & 215.0 & $-$18.4\% \\
\textbf{MAESTRO-36k $\to$ poetry $\to$ prose} &
  \textbf{196.3} & \textbf{$-$25.4\%} \\
\bottomrule
\end{tabular}
\end{table}

Two observations follow.
First, \textbf{adding the poetry phase at $d\!=\!32$ improves
epoch-2 PPL by an additional $-8.7\%$} over the music-only transfer
(215.0 $\to$ 196.3), confirming at a second scale the additive
benefit of the poetry phase observed in Phase 2 at $d\!=\!16$.
Second, \textbf{the full-pipeline advantage over random
initialization grows with scale}: $-17.5\%$ at $d\!=\!16$
(Phase 2, 5 seeds) to $-25.4\%$ at $d\!=\!32$ (this subsection,
seed 42).
This trajectory is consistent with Phase 3's capacity-saturation
finding---larger models extract more value from the pipeline---and
with the persistent plateau-level gap at $d\!=\!64$
(\S\ref{sec:convergence}).
The $d\!=\!32$ result is single-seed and should be interpreted
with the same caution as the Phase 3 table; we report it here to
establish the full-pipeline trajectory across all three scales
rather than as a statistically validated claim.

\subsection{Convergence Test: Does the Gap Survive Long-Term Training?}
\label{sec:convergence}

Phases 1--3 evaluate language learning at a fixed three-epoch horizon.
A natural concern is that the pipeline's advantage reflects a
\emph{head start} that vanishes once the random baseline trains
long enough.
To address this, we train both conditions---random initialization
and the full developmental pipeline---until plateau
(defined as $<2.5\%$ improvement per epoch, with a minimum of
2 epochs before triggering).
Each scale uses the best music data size identified in Phase 3:
MAESTRO-12k for $d\!=\!16$ and MAESTRO-36k for $d\!=\!64$.
To establish statistical reliability at the larger scale,
we run the $d\!=\!64$ convergence test across five seeds
(42, 123, 456, 789, 1024), matching the Phase~2 validation protocol.

\paragraph{Single-seed pilot ($d\!=\!16$).}
At $d\!=\!16$ (seed 42), the random baseline converges to PPL 346.8
after 7 epochs, while the pipeline reaches 322.5 in just 5 epochs---a
7.0\% gap.
This initial result motivated the multi-seed validation at $d\!=\!64$.

\paragraph{Multi-seed validation ($d\!=\!64$).}
Table~\ref{tab:convergence} reports the per-seed results.

\begin{table}[h]
\centering
\caption{Convergence test at $d\!=\!64$: training to plateau
(5 seeds). The pipeline uses MAESTRO-36k + 3 epochs of poetry
before prose.
Gap is the percentage improvement of pipeline over random
at convergence.}
\label{tab:convergence}
\begin{tabular}{@{}rrrrr@{}}
\toprule
Seed & Random PPL (ep) & Pipeline PPL (ep) & Gap \\
\midrule
42   & 122.0 (8) & 112.9 (7) & $-$7.5\% \\
123  & 117.9 (9) & 116.1 (6) & $-$1.5\% \\
456  & 118.3 (9) & 114.9 (6) & $-$2.9\% \\
789  & 122.3 (8) & 111.4 (7) & $-$8.9\% \\
1024 & 117.8 (9) & 109.8 (8) & $-$6.8\% \\
\midrule
Mean $\pm$ std & $119.7 \pm 2.3$ & $\mathbf{113.0 \pm 2.6}$ & $-$\textbf{5.5\%} \\
\bottomrule
\end{tabular}
\end{table}

The pipeline outperforms the random baseline in all five seeds,
with a mean gap of 5.5\% ($t = 3.90$, $p = 0.017$, paired $t$-test,
$n = 5$).
Three results emerge.

First, \textbf{the gap does not close}.
Across all five seeds, the random baseline cannot reach the
pipeline's converged perplexity even with additional epochs
(8--9 epochs for random vs.\ 6--8 for pipeline).
This rules out the ``catch-up'' hypothesis: the pipeline's
advantage is not merely an acceleration effect but reflects
a genuinely lower loss basin.

Second, \textbf{the gap varies across seeds but is consistently
positive}.
Individual gaps range from 1.5\% to 8.9\%, reflecting
seed-dependent variation in how effectively the transferred
structures are exploited during language learning.
Despite this variation, the pipeline wins in every seed,
and the mean effect is statistically significant.

Third, \textbf{the pipeline converges faster}.
In every seed, the pipeline reaches plateau in fewer epochs
than the random baseline (6--8 vs.\ 8--9).
The developmental pipeline thus provides both a better destination
and a faster path to reach it.

Note that the convergence gap (5.5\% at $d\!=\!64$) is smaller than
the three-epoch gap reported in Phase 2 (11.8--17.5\%).
This is expected: additional training epochs benefit both conditions,
but the random baseline has more room for improvement from its
higher starting point.
The persistent gap at plateau represents the
\emph{irreducible} advantage of musical pre-training---structural
knowledge that cannot be recovered by training longer on text alone.

\section{Analysis}
\label{sec:analysis}

The preceding results establish \emph{what} happens;
here we interpret \emph{why}, organized around six themes.

\subsection{Why Does Music Transfer to Language?}

The transferred attention weights carry computational
structures---long-range dependency tracking, hierarchical pattern
recognition---that directly benefit language processing.
This is consistent with \citet{lee2026nca}'s finding that
attention weights are the most transferable component,
and extends it by showing that music provides a particularly
effective source domain.
Table~\ref{tab:music-language} summarizes the structural parallels
that may explain this transfer.

\begin{table}[h]
\centering
\caption{Structural parallels between music and language.}
\label{tab:music-language}
\begin{tabular}{@{}lll@{}}
\toprule
Structure & Music & Language \\
\midrule
Hierarchy & note $\to$ phrase $\to$ section & word $\to$ clause $\to$ paragraph \\
Long-range dep. & theme $\to$ development $\to$ recap. & subject--verb agreement \\
Expectation & dissonance $\to$ resolution & garden-path correction \\
Directionality & tension $\to$ resolution & given $\to$ new information \\
\bottomrule
\end{tabular}
\end{table}

\subsection{Quality as Efficiency}

MAESTRO-12k and Synth-36k reach statistically indistinguishable
ceilings (Section~\ref{sec:transfer}), so the quality advantage
of real music is best characterized as an \emph{efficiency}
advantage: denser structural information per data point,
enabling the model to saturate sooner.
Synthetic data, being structurally simpler, requires more
examples to extract the same amount of transferable structure.

\subsection{Orthogonal Contributions Explain Additivity}

Music and poetry improve \emph{different model components},
and this separation is a direct consequence of the transfer
design (Section~\ref{sec:pipeline}).
When transitioning from music to poetry, only
vocabulary-independent layers---attention, feedforward, and
layer normalization weights---are transferred;
the token embedding and language model head are reinitialized
because music tokens and language tokens have no correspondence.
Consequently, any benefit of music pre-training is confined
\emph{by construction} to internal computation
(attention + FFN), and cannot reside in embeddings.

The epoch-0 perplexities confirm this decomposition.
After music-only transfer (MAESTRO-12k $\to$ Prose),
the initial perplexity is ${\sim}500$---still high because
the reinitialized embeddings carry no language information---yet
subsequent learning proceeds significantly faster than the
random baseline ($-11.8\%$ at epoch 2).
This pattern is the signature of improved internal computation:
the model does not \emph{start} better, but \emph{learns} faster.
After the poetry phase (MAESTRO $\to$ Poetry $\to$ Prose),
the initial perplexity drops to $416 \pm 4$, already below
the random baseline's \emph{converged} value ($423.0$).
Since the internal weights are unchanged between the poetry
and prose phases (same vocabulary, full model continues),
this additional drop is attributable to embedding calibration.

The near-additivity of these effects further supports
orthogonality: music alone improves perplexity by 11.8\%,
the poetry phase adds a further 5.7 percentage points
(17.5\% total), and this increment is consistent across
all five seeds.
If music and poetry competed for the same model capacity,
we would expect sub-additive gains; instead, the effects
compose because they target non-overlapping parameters.

\subsection{A Developmental Ordering}

The optimal training order---music $\to$ poetry $\to$ prose---mirrors
a well-documented pattern in human infant development:
sensitivity to pre-linguistic regularities (rhythm, prosody;
0--6 months) $\to$ phonemic discrimination (6--12 months) $\to$
vocabulary and grammar (12+ months) \citep{kuhl2004early}.
We suggest this is not coincidence: acquiring a primitive sense
of regularity before linguistic content is a fundamentally
efficient strategy, and biological development has converged on
this solution through evolutionary pressure.
Both systems face the same computational problem---bootstrapping
language comprehension from sub-linguistic pattern recognition---and
both benefit from the same curriculum: regularity first, language second.
This principle generalizes curriculum learning
\citep{bengio2009curriculum} by grounding the ordering not in
task difficulty but in the progression from non-linguistic
to linguistic structure.

\subsection{Capacity-Dependent Data Curation}

The monotonic shift of optimal data volume with model capacity
(Section~\ref{sec:scale}) has a practical implication:
\textbf{pre-training data volume should be calibrated to model size}.
Unlike language pre-training, where more data is almost always
beneficial \citep{hoffmann2022training}, music pre-pre-training
has a \emph{capacity-dependent} optimum---the purpose is not
to acquire domain knowledge (which scales with data) but to
establish computational structures (which saturate once the
model has learned the relevant patterns).

This extends \citeauthor{lee2026nca}'s observation that optimal
data \emph{complexity} varies by target domain.
Our result adds a second axis: optimal \emph{volume} varies by
model capacity, defining a two-dimensional design space that
should be navigated as a function of both the target domain and
the model's absorptive capacity.

\subsection{Irreducible Transfer}

The convergence tests (Section~\ref{sec:convergence}) show that
the pipeline's advantage is not a transient head start but an
\emph{irreducible} benefit.
At $d\!=\!64$, multi-seed validation confirms a mean 5.5\% gap
at plateau ($p = 0.017$), with the pipeline winning in all five seeds.
This transforms the interpretation of the three-epoch results.
The 11.8--17.5\% improvements are the early manifestation of a
persistent advantage that narrows as both conditions approach
their respective ceilings, but never vanishes.
Music pre-training provides structural knowledge that
\emph{cannot be recovered} by training longer on text alone.

\section{Limitations}
\label{sec:limitations}

Our study has several limitations.
First, and most importantly, the scale of our experiments
is small: Phase 3 extends our findings only to $d\!=\!64$
(${\sim}400$K parameters), several orders of magnitude below
modern language model pre-training.
Whether the observed transfer effects---their magnitude, their
orthogonality between music and poetry, and the capacity--data
scaling trend---persist at millions or billions of parameters
is an open empirical question that the present experiments
cannot settle; stronger conclusions will require scaling up.
Second, we evaluate only on English with classical Western music;
generalization to other languages and musical traditions is unknown.
Third, we report only perplexity; downstream task evaluation
would strengthen the findings.
Fourth, the music-to-language transfer requires vocabulary
reinitialization, which discards information;
more sophisticated transfer methods
(\eg, shared subword-MIDI vocabularies) may improve results.
Fifth, while Phase 2 validates the main findings across five
random seeds, the music checkpoints themselves use a single seed
(seed 42), and Phase 3 (the scale experiment), including the
$d\!=\!32$ full-pipeline result in \S\ref{sec:phase3-full}, uses a
single seed throughout; multi-seed replication of the scaling
results would strengthen the conclusions.
Finally, the $d\!=\!16$ convergence test uses a single seed;
while the $d\!=\!64$ convergence test is validated across
five seeds ($p = 0.017$), multi-seed replication at $d\!=\!16$
would further strengthen the cross-scale comparison.

\section{Conclusion}
\label{sec:conclusion}

We have demonstrated that music pre-pre-training accelerates
language learning in Transformers, with the full developmental
pipeline achieving a 17.5\% perplexity improvement ($p < 0.001$).
Six findings emerge from our experiments:
(1) real music by master composers provides a highly effective
pre-training substrate, achieving the same transfer as synthetic
patterns with one-third the data;
(2) a developmental pipeline---music $\to$ poetry $\to$ prose---yields
additive improvements that no single phase can match;
(3) music and poetry improve orthogonal model components---internal
computation and embeddings, respectively;
(4) the poetry phase is so effective that the model enters prose
training with perplexity already below the untrained baseline's
final converged value;
(5) optimal pre-training data volume scales with model capacity,
with the advantage of larger datasets growing monotonically
from $d\!=\!16$ through $d\!=\!64$; and
(6) the pipeline's advantage persists at convergence---multi-seed
validation at $d\!=\!64$ confirms a 5.5\% gap at plateau
($p = 0.017$), with the pipeline winning in every seed,
confirming that the transferred structures provide
an irreducible benefit rather than a transient head start.
Together, these results suggest a developmental pipeline for
language models---\emph{listen, chant, then read}---grounded in the
same progression that characterizes human language acquisition,
with data volume calibrated to the model's absorptive capacity.

More broadly, our findings suggest that the quality of
pre-training data---specifically, whether it is a product of
skilled human creative activity---matters for efficiency:
real music encodes more transferable structure per data point
than synthetic alternatives.
The capacity-dependent scaling of optimal data volume further
implies that pre-training data curation should be treated as
a joint function of data quality, data quantity, and model size.

We emphasize that all of these results are obtained at small
scale (at most $d\!=\!64$, ${\sim}400$K parameters) and on a
narrow empirical footprint (English prose, Western classical
piano, REMI-style MIDI tokenization).
They should therefore be read as evidence that the developmental
pipeline is \emph{worth scaling up and stress-testing}, not as
a demonstration that the effect holds at the scale of modern
language models.
Establishing whether the music $\to$ poetry $\to$ prose pipeline
produces comparable relative gains---under FLOPs-matched
comparisons, at parameter counts several orders of magnitude
larger, across multiple natural languages and musical traditions,
and on downstream tasks beyond perplexity---will require
substantially larger experiments than those reported here,
and is the main direction in which the present study needs to be
extended before stronger conclusions can be drawn.

\section{Future Work}
\label{sec:future}

Several directions merit investigation.
First, validating the developmental pipeline at larger scales
($d\!=\!128$, $d\!=\!256$) with a \emph{FLOPs-matched} comparison
against Wikipedia-only training would establish whether the
pipeline's efficiency advantage holds when computational budgets
are equalized at practical model sizes.
Second, the structural similarity between our capacity-constrained
setting and low-rank adaptation (LoRA) suggests that
\emph{developmental initialization}---warming up LoRA's
low-rank subspace with structured data before domain
fine-tuning---may improve fine-tuning efficiency, particularly
at low ranks ($r = 4$--$8$) where the subspace direction matters most.
Third, extending the pipeline to non-Western music and
non-English languages would test whether the transfer relies on
universal structural properties or culture-specific patterns.
Finally, mapping the full interaction surface---music data volume
$\times$ poetry data volume $\times$ model capacity---would enable
principled curriculum design for the developmental pipeline
at any target scale.


\bibliographystyle{plainnat}
\bibliography{references}

\end{document}